# A Machine Learning-based Algorithm for Automated Detection of Frequency-based Events in Recorded Time Series of Sensor Data

**Bahareh Medghalchi** [a b,*], **Andreas Vogel** [b]

a. Volkswagen AG, Berliner Ring 2, 38436 Wolfsburg, Germany
b. Institute of High Performance Computing, Ruhr University of Bochum, Universitätsstraße 150, 44801 Bochum, Germany

*E-Mail Addresses: bahareh.medghalchi@volkswagen.de (B. Medghalchi), a.vogel@rub.de (A. Vogel)

**Abstract:**

Automated event detection has emerged as one of the fundamental practices to monitor the behavior of technical systems by means of sensor data. In the automotive industry, these methods are in high demand for tracing events in time series data. For assessing the active vehicle safety systems, a diverse range of driving scenarios is conducted. These scenarios involve the recording of the vehicle's behavior using external sensors, enabling the evaluation of operational performance. In such setting, automated detection methods not only accelerate but also standardize and objectify the evaluation by avoiding subjective, human-based appraisals in the data inspection. This work proposes a novel event detection method that allows to identify frequency-based events in time series data. To this aim, the time series data is mapped to representations in the time-frequency domain, known as scalograms. After filtering scalograms to enhance relevant parts of the signal, an object detection model is trained to detect the desired event objects in the scalograms. For the analysis of unseen time series data, events can be detected in their scalograms with the trained object detection model and are thereafter mapped back to the time series data to mark the corresponding time interval. The algorithm, evaluated on unseen datasets, achieves a precision rate of 0.97 in event detection, providing sharp time interval boundaries whose accurate indication by human visual inspection is challenging. Incorporating this method into the vehicle development process enhances the accuracy and reliability of event detection, which holds major importance for rapid testing analysis.

## 1 Introduction:

The safety systems in vehicles are generally divided into two categories, namely the Passive Vehicle Safety (PVS) systems and the Active Vehicle Safety (AVS) systems. PVS systems mitigate the severity of injuries and damages of crashes on the occupants and other road users whereas AVS systems prevent the occurrence of accidents either through warning the driver and calling the conductor's attention to the dangerous situation or by intervening in driving and keeping the vehicle away from the risk of accident. (Jeong and Oh, 2017) To develop robust AVS systems in the vehicles, their operations must be controlled and monitored with the help of external testing facilities to assess the efficiency of the AVS systems. To achieve this, various real world driving scenarios are simulated and externally installed sensors measure and record various parameters of the vehicle, such as torsional moment on the steering column, steering robot angle, rotational speed of the steering wheel and torque applied on the steering wheel. These sensors should allow to identify one or multiple triggered AVS functions. Thus, the recorded data is inspected to ensure whether or not a particular AVS function has been activated promptly and on time during the test maneuver. Identifying the drive events that occurred throughout the AVS tests based on the recorded time series data is one of the challenging steps for data analysts in the automotive industry.

In the light of the considerable amount of time series data captured during AVS tests, it is imperative to constantly monitor the recorded data with substantial expertise as well as the requisite knowledge and precision to interpret the test procedure based on the acquired data. Manually monitoring the collected data is a laborious and time-intensive task, thus prompting the necessity for an automated methodology that monitors time series data and tracks relevant events with high efficiency and a precise time interval for the detected events. (Alvarez-Coello et al., 2019) Identifying the points of interest with specific significance for the user in the time series data is referred as event detection, with the event being defined by the user in the majority of occasions. (Gensler and Sick, 2018) In the field of signal processing, it is a common practice to monitor the time series data for tracking the meaningful changes which characterize the occurrence of an event at a certain moment. Therefore, event detection is a widely considered topic in applications of sensor data analysis. (Lima et al., 2022)

The objective of this study is to formulate a methodology that can reliably discern the incidence of an event, encompassing the exact temporal boundaries of its initiation and termination, through the utilization of sensor data. Our novel monitoring approach involves transforming the time series data into visual representations in time-frequency domain employing scalograms, which are then filtered to accentuate the desired event and suppress the irrelevant parts of the scalograms. Next, an object detection network is used to trace the objects representing the intended event in the scalograms. Subsequently, the traced objects in the scalograms are mapped back to the time series data. To demonstrate the practical relevance of our method, we apply it to track the activation of one of the AVS systems in vehicles, namely Lane Departure Warning (LDW), using sensor data. Specifically, we aim to detect the vibrations generated on the steering wheel due to lane departure.

The main contributions of this work are given as follows:

- we propose to adapt the concept of object detection to event detection in time series data,
- we present a workflow including data transformation to scalograms, filtering the scalograms , and training a YOLO object detection network which allows to detect frequency-based events in time series data,
- we take advantage of the precision of object detection networks in object localizing in order to identify precise time intervals of the occurred events,
- we quantify and show the effectiveness of the proposed method by applying it to AVS data sets.

To the best of our knowledge, a comparable workflow for event detection within the domain of signal processing has not yet been proposed.

The remainder of this paper is structured as follows: Section 2 provides an overview of related works. Section 3 introduces a specific type of LDW function whose operation is aimed to be monitored through event detection and gives a brief overview about data acquisition procedure. In Section 4, a comprehensive methodology ranging from data preprocessing to training algorithm is demonstrated. Section 5 presents the results and discussion and in the end, Section 6 summarizes the method, the major contribution of this research, and future extension.

## 2    Related Work

For identifying driving events based on the time series data gathered via sensors, mainly three approaches have been introduced in the literature: threshold-based methods, anomaly detection methods and Machine Learning (ML) classifiers.( Gatteschi et al., 2022) Threshold-based methods are the simplest approaches that extract fixed limits to detect the events. These approaches are not optimal solutions in identifying the most complex real-world events because different driving types, road conditions, different vehicles, test locations and sensor characteristics can affect the thresholds; in addition, extracting a threshold which can remain robust under all these varying conditions can be challenging.(Carlos et al., 2018; Eboli et al., 2016, 2017). In the case of using anomaly and outlier detection methods, the outlying data does not have to be specifically defined in advance; however, any deviation from the majority of samples can be considered as anomaly and defect. (Costa et al., 2015) Moreover, identifying particular samples as event among numerous normal samples is more challenging than detecting anomalies and outliers. (Benkő et al., 2022) As an example for determining the status of the system and checking whether a specific event has occurred using anomaly detection methods, the authors of (Oehmcke et al., 2015) have made use of outlier scores to measure the difference between the given inference samples and the labeled class of the samples. These approaches are unsupervised ML methods which could be advantageous when only specific types of classes exist in the data and any deviation from those classes is intended to be detected. However, the process of detecting an event varies significantly from that of anomaly detection as in this method, a certain event does not necessarily differ from the majority of the data sets.( Benkő et al., 2022) A considerable number of recent studies in the field of event detection have focused on using ML based single-, binary- or multi-class classifiers. For example, the methodology proposed in (Zylius, 2017) has extracted numerical features from recorded accelerometer data and after identifying the best features, the authors classified the data into safe and unsafe driving statuses using the Random Forest (RF) classification method. The method has achieved an accuracy of 95.5% in classifying the time series data into two classes. In addition, the authors of (Bejani and Ghatee, 2018) have employed multiple classifiers, including Support Vector Machine (SVM), K-Nearest Neighbors (K-NN), Decision Tree (DT), and Multi-Layer Perceptron (MLP) techniques, to classify driving data into safe and unsafe maneuvers. Apart from studies such as (Ahmed and Begum 2020; Eftekhari and Ghatee, 2019; Júnior et al., 2017; Li et al., 2017) which have used the binary classification methods to detect specific maneuvers throughout the drive, a great number of studies have employed ML classifiers to identify multiple driving maneuvers and classify the recorded data into multiple classes. The authors of (Eftekhari and Ghatee, 2018; Yuksel and Atmaca, 2021) have compared the efficiency of different classifiers such as Bayesian Networks (BNs), MLP, DT, SVM and Naïve Bayes(NB) in classifying the driving maneuvers. One Class Classifiers (OCC) are also regarded as highly favored concepts in the field of detecting driving events based on the recorded time series data. In (Martínez-Rego et al., 2016), a fault detection method based on OCC concept has been proposed whereby a single class has been defined and trained for the normal driving maneuvers and the data deviating significantly from this class are disregarded.

One of the main challenges in using the classifiers for detecting the events along the time series data is finding an appropriate method to take samples of the data. Sliding window is one of the widely used techniques in taking samples of the data via dividing the time series data into frames of the fixed length according to a specific frame and step size. (Barandas et al., 2020) For instance, the study described in (Gensler and Sick, 2018) has created windows of predefined length and step along the time series data and after extracting the numerical features from each window, the windows overlapping the desired events have been labeled as event containing windows and used to train a multivariate Gaussian classifier. For evaluating the proposed method, the occurrence of the event throughout an unseen window of time series was checked by measuring the difference between the features of the unseen window and the centroid of the trained class of event containing windows. The work proposed by (Lattanzi and Freschi, 2021) outlines an approach for recognizing safe and unsafe driving behaviors through the collection of time series data from in-vehicle sensors. The methodology involves windowing the collected data and extracting a set of descriptive numerical features for each window. The resulting windows are then classified using SVM and feed-forward neural networks. Research in (Shahverdy et al., 2020) presents a technique for categorizing windows of collected data from car sensors, which includes acceleration, gravity, speed, and Engine Revolutions Per Minute (RPM). The data is transformed into images using a recurrence plot method and then fed into a Convolutional Neural Networks (CNN) for classification into five class of driving styles.

In addition to the approaches introduced for event detection in the field of automotive engineering, many studies in other fields of science have made use of ML-classifiers for identifying events in time series data as well. For medical applications such as diagnosing irregularities in the Electrocardiogram (ECG) signals, the author of (Marzog and Abd, 2022) has proposed a method which extracts numerical features from ECG time series data and then, categorizes the data using different classification methods from the simplest ones such as SVM and DT to the most complicated classifiers such as Deep Learning (DL) methods. In (Kim et al., 2018), the authors have used CNN to classify the recorded audio and radar data in order to distinguish five different events. The authors of (Bajaj et al., 2019; Kaya, 2020) have proposed an event detection method for Electroencephalogram (EEG) time series data, in which after extracting the numerical features from EEG data, KNN classifiers have been deployed to classify the data.

To develop an advanced event detector in the field of signal processing, a delicate feature extraction procedure is necessary to effectively highlight the event in time series data. (Benkő et al., 2022) Calculating Continues Wavelet Transformation (CWT) from time series data is one of the efficient feature extraction methods used extensively in many studies. The authors of (Halberstadt, 2020) have transformed the windows of time series data into the visual representations of scalograms and then have classified the scalogram into two classes of containing and non-containing a certain event. The authors of (Er and Aydilek, 2019) addressed a different task, i.e., recognizing the emotional content of the recorded musical pieces. They have generated specific visual representations from the music records, namely chroma spectrogram. Chroma spectrograms depict the energy distribution of music notes over time. The authors have used chroma spectrograms to train classifiers to recognize the emotional content of the music pieces. Similarly, in (Almanza-Conejo et al., 2022), a ML-based classification method has been proposed for identifying emotions based on EEG signals. After transforming the windows of discrete time series data into scalograms, CNN is used to extract features from the images.

Subsequently, the scalograms are classified into multiple classes. The authors of (Copiaco et al., 2019) have compared the performance of different pre-trained DL networks combined with the Linear SVM in classifying the scalograms extracted from the audio data. As the introduced studies have employed classifiers for event detection, necessitating the sequential process of initially windowing and subsequently classifying the time series frames, they are still subject to the limitations associated with windowing the temporal data, even despite leveraging the advantages offered by generating scalograms from the time series data.

While a substantial number of studies have employed approaches that detect events using ML classifiers, an optimal solution has not yet been identified (Gatteschi et al., 2022). Since a large number of labeled datasets is generally required to train classification methods, the scarcity of data available in most studies is one reason why classification methods may not be an efficient approach for event detection. The second setback for training an optimal classifier is that the performance of the classifiers could be easily affected by different factors, such as the type of vehicle under test, measuring device and different driving conditions when detecting driving events. (Elassad et al., 2020) The third reason why classifiers are deemed unsuitable for event detection is their tendency to capture imprecise temporal intervals for the events. Obtaining the precise time interval during which an event occurred is of paramount importance, particularly in this study. As previously mentioned, one of the main challenges of using classifiers for detecting events in time series data is finding an appropriate data sampling method. To address this issue, windowing has been used in numerous studies. This involves dividing the time series data into fixed-size windows and extracting features from each window frame. The classifier then categorizes the entire window into a specific class. However, due to the fixed frame and step sizes of these windows, it is unlikely for the time interval of an occurred event to perfectly align with an integer number of consecutive windows. As a result, the classifiers often fail to detect the precise time interval for an event. (Alvarez-Coello et al., 2019) One more drawback of dividing the entire data into windows is that, by dividing the data into windows of the same size and carrying out feature extraction on each window, the computational cost of the ML-based classifiers gets one order larger than the computational cost of the classifier itself. (Gatteschi et al., 2022) Table I provides an overview of previous studies within the domain of signal processing and their alignment with the prerequisites of the event detection methodology. Given that none of the contemporary inquiries have satisfied all requirements, this manuscript has formulated an innovative methodology capable of satisfying the entirety of these exigencies.

*Table I: Previous studies within the domain of signal processing and their alignment with the prerequisites of the event detection methods. ( $-$: not fulfilled, $\checkmark$: fulfilled )*

| Related works in literature: | The principle method used: | Requirements to be fulfilled | | |
|---|---|---|---|---|
| | | Discerning the incidence of the event | Providing exact time interval boundaries | Remaining robust by influential and varying factors and the driving conditions |
| Carlos et al., 2018 | threshold-based | √ | √ | — |
| Eboli et al., 2016 | threshold-based | √ | √ | — |
| Eboli et al., 2017 | threshold-based | √ | √ | — |
| Oehmcke et al., 2015 | outlier scores | — | √ | √ |
| Zylius, 2017 | binary classifiers | √ | — | √ |
| Bejani and Ghatee, 2018 | binary classifiers | √ | — | √ |
| Ahmed and Begum, 2020 | binary classifiers | √ | — | √ |
| Eftekhari and Ghatee, 2019 | binary classifiers | √ | — | √ |
| Júnior et al., 2017 | binary classifiers | √ | — | √ |
| Li et al., 2017 | binary classifiers | √ | — | √ |
| Eftekhari and Ghatee, 2018 | multi-class classifiers | √ | — | — |
| Yuksel and Atmaca, 2021 | multi-class classifiers | √ | — | — |
| Martínez-Rego et al., 2016 | one class classifier | √ | — | √ |
| Gensler and Sick, 2018 | multi-class classifiers | √ | — | — |
| Lattanzi and Freschi, 2021 | binary classifiers | √ | — | √ |
| Shahverdy et al., 2020 | multi-class classifiers | √ | — | — |
| Marzog and Abd, 2022 | multi-class classifiers | √ | — | — |
| Kim et al., 2018 | multi-class classifiers | √ | — | — |
| Bajaj et al., 2019 | multi-class classifiers | √ | — | — |
| Kaya, 2020 | multi-class classifiers | √ | — | — |
| Halberstadt, 2020 | binary classifiers | √ | — | √ |
| Er and Aydilek, 2019 | multi-class classifiers | √ | — | — |
| Almanza-Conejo et al., 2022 | multi-class classifiers | √ | — | — |
| Copiaco et al., 2019 | multi-class classifiers | √ | — | — |

## 3 Active Vehicle Safety (AVS) Tests and Sensor Data Acquisition

Automated safety reactions of a vehicle are one of the factors which are under constant development and modernization to ease the driving experience. Lane Support System (LSS) is a term used for a group of AVS operations which intend to keep the vehicle within the lane by detecting the lane markings on the road. (Pappalardo et al., 2022) LSS systems include three operational levels referred to as Emergency Lane Keeping (ELK), Lane Keep Assist (LKA) and Lane Departure Warning (LDW). While LKA tries to keep the vehicle within the lines, ELK intervenes the driving behavior when a critical situation is detected; meanwhile, LDW emits a warning such as vibration or a signal tone in the case of any deviation from the lanes. (Dollorenzo et al., 2022) Recently, one of the most common forms of the LDW warning is issuing vibration on the steering wheel that alerts the driver about the lane departure, highlighting the significance of timely warning.

For assessing the operation of the AVS systems in different driving conditions, vehicle testing institutes such as European New Car Assessment Programme (Euro NCAP) carry out several AVS tests. By simulating various test scenarios and depending on the driving experience of the test driver, the performance of the vehicle's AVS system is evaluated. One aspect in assessing the performance of LDW is checking whether a well-timed warning is produced or not, which is one of the standard guidelines, according to which Euro NCAP rates the vehicles. A protocol of test scenarios which should be carried out to assess the operation of the LSS systems is called LSS tests.

By convention, evaluating the LDW is carried out according to questionnaire-based surveys in which the test driver is asked to report the instance in which a vibration on the steering wheel has been sensed. A drawback of the self-reported controlling approaches is that they are highly affected by social desirability bias and human errors, and questionnaires are usually based on the subjective views of the drivers rather than the actual performance of the systems involved in the AVS test (Gatteschi et al., 2022). Furthermore, questionnaires are mainly appropriate for qualitative analyses and not for the assessments where quantitative parameters should be evaluated precisely. Therefore, an automated controlling method that detects the occurrence of an event along with its exact start and ending time points based on the recorded data would ease the surveillance process.

For the proposed method in this work, the time series data is collected by sensors which are externally installed to the vehicle and measure various parameters of the vehicle with varying sampling rates. All sensors and measuring equipment are connected to a single data acquisition source which synchronizes the entire recorded data.

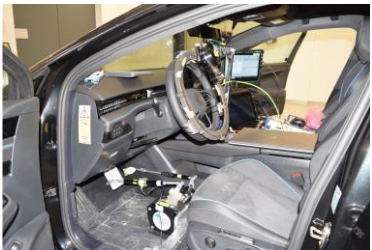 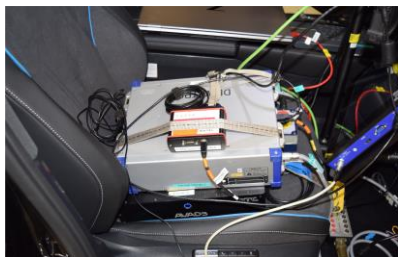 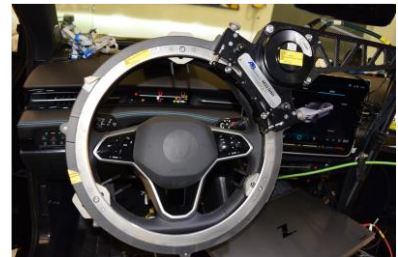

Fig. 1(a). AVS Test Equipment

Fig. 1(b). Data Measuring and Acquisition equipment

Fig 1(c). Steering Robot

To be more precise, for the specified starting time, end time and the given frequency, the data acquisition source interpolates the values for the time points which have not been measured directly and skips the values that have been measured in frequencies higher than the required ones. Figures 1.1 to 1.3 show a data acquisition system configured to record the data from the sensors in the LSS test. Through empirical analysis of the data collected from such LSS test setups, it was ascertained that the sensor measuring the moment on the steering column detects vibrations associated with LDW functionality. Therefore, monitoring the corresponding recorded data can be used to find out whether the LDW warning has been issued and if this is the case, at what exact time interval LDW vibration was activated.

The vibration issued on the steering wheel is also observable in the data gathered from the Controller Area Network (CAN-Bus). The CAN-Bus is a standard communication unit inside modern vehicles through which different electronic control units are connected together. (Fugiglando et al., 2019) However, since the external impact of the AVS systems on the occupants is controlled throughout the AVS tests, the internal CAN-Bus data are not used for the Euro NCAP audition. In addition, it is time-consuming to access CAN-Bus data since the CAN-Bus sensors are installed internally in the vehicle and access is typically restricted, requiring specialized hardware and setup. Thus, the functionality of the AVS systems is analyzed through sensors which are installed externally on the vehicle and in the case of LDW vibration, by means of the sensors implemented on the steering robot. Notwithstanding the inadmissibility for auditions, the recorded CAN-Bus sensor data can serve as a reliable reference to evaluate and validate the analysis method proposed in this work. It will serve as the ground truth indicating the accurate time interval for the issued vibration on the steering wheel. Figure 2 depicts recorded data from the external sensor measuring steering column moment and the corresponding CAN-Bus sensor measuring steering wheel angle. LDW activation intervals are marked in both datasets. While the LDW events are clearly visible in the CAN-Bus data, their identification is challenging in the steering column moment data due to background vibrations. The proposed algorithm aims to identify and detect these intervals solely based on the steering column moment sensor data.

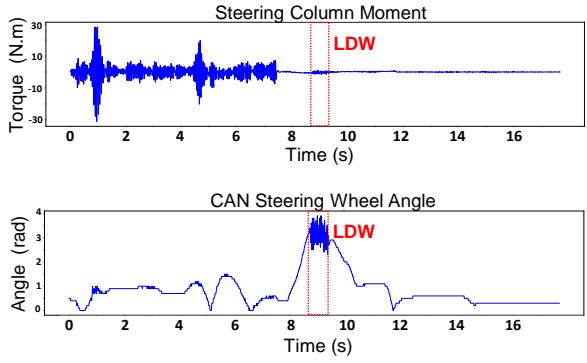

Fig. 2(a).

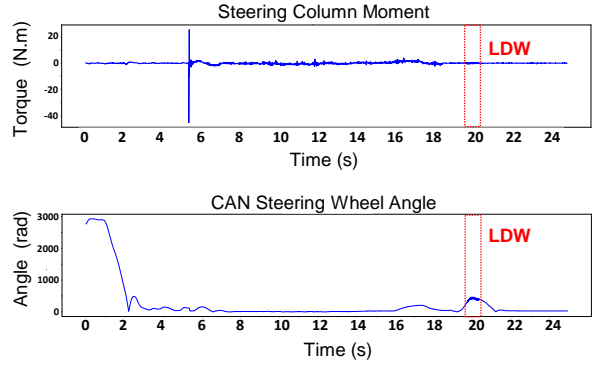

Fig. 2(d).

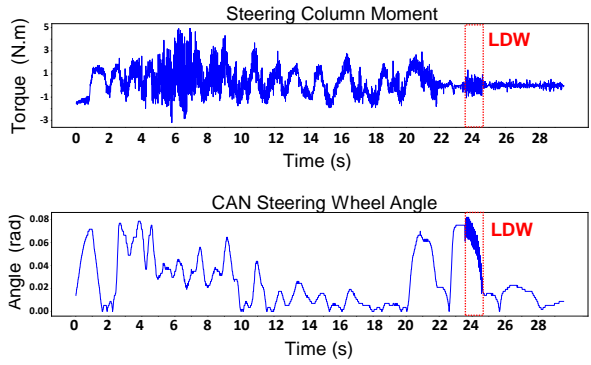

Fig. 2(b).

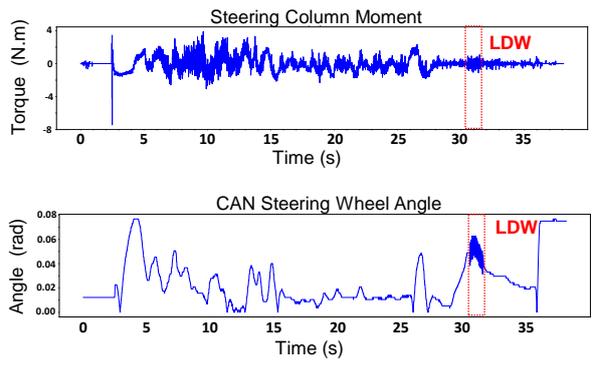

Fig. 2(e).

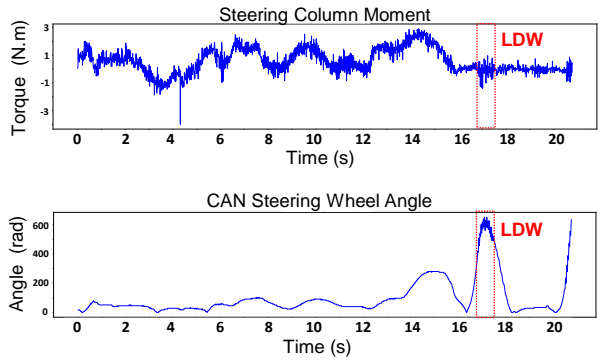

Fig. 2(c).

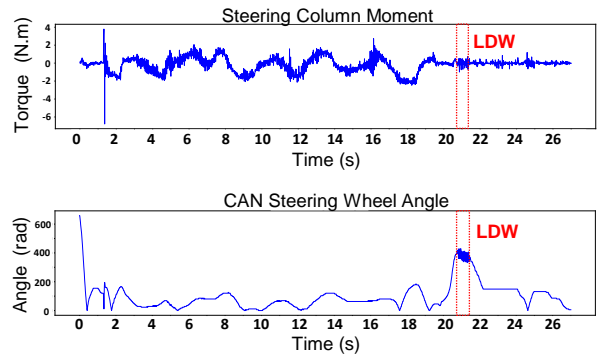

Fig. 2(f).

Fig. 2. The recorded time series data from the externally installed sensor measuring the moment on the steering column (top) and the corresponding CAN-Bus sensor (bottom) is shown for exemplary LSS tests.

## 4 Event Detection Methodology

Machine Learning (ML) methods are powerful tools in automatically identifying events based on recorded time series data. Compared to traditional event detection methods in the field of signal processing, ML approaches are able to automatically track a certain event from a large set of recorded data and reduce the dependency on expert experience and knowledge as well as resolving the uncertainty and subjectivity originating from manual monitoring. (Gatteschi et al., 2022) In the following, we describe the proposed method for event detection in time series data.

### 4.1 Data Preprocessing

#### 4.1.1 Continuous Wavelet Transform (CWT)

An accurate condition monitoring algorithm requires features of high quality extracted from time series data. The Continuous Wavelet Transform (CWT) is a powerful feature extraction tool that enables the characterization of the mapping relationship between system's status and the recorded data. This is achieved by decomposing time series data into the superpositions of base wavelets of multiple scales. As a result, CWT compensates for the loss of high frequency components of the time series data, making it an effective method for analyzing complex signals. Moreover, the magnitude of the CWT drawn as RGB color scale improves the readability and comprehensibility of time series data and can be used as input for DL networks, capable of time-localizing short time events as well as low frequency events. (Zhao et al., 2023) Although many wavelet-based event detection techniques have been developed for stationary time series data such as (Abu-Elanien and Salama, 2009; Costa, 2014; Dwivedi and Singh, 2010), time series data recorded by sensors throughout AVS test have nonstationary properties, rendering the event detection much more complicated. However, scalograms are still one of the ideal tools in detecting phenomena of short duration for nonstationary time series data owing to their spectro-temporal characteristics which take both the time and frequency aspects of times series data into account. (Kim et al. 2017) This is beneficial in projecting the properties of the constant events in nonstationary time series data with minimum information loss. (Copiaco et al., 2019)

In order to generate scalograms from time series data, a base wavelet is required which is a waveform of limited duration with an average of zero. Different kinds of base wavelet can be used such as Gabor wavelet, Morse and Bump wavelets and choosing the right base wavelet plays a significant role in mapping the time series data.(Costa, 2014) For this study, we employ Gabor wavelet as base wavelets which are analytic wavelets with complex values in the time domain and have equal variance in time and frequency. The Gabor wavelet is defined as

$$\psi(t) = exp(j\omega_0 t)\ exp\ (-t^2/2) \qquad (1)$$

where $\omega_0$ is the frequency of the wavelet. The CWT of a time series data of $f(t)$ is then calculated by

$$CWT(a,b) = 1/\sqrt{a} \int_{-\infty}^{+\infty} f(t).\psi((t-b)/a)\ dt, \quad a,b\ \epsilon\ \mathbb{R}, \qquad (2)$$

where $a$ is the scale of the wavelet, which either compresses the wavelet or stretches it out and $b$ is the position of the wavelet. The value of the CWT is computed for different values of scale $a$ and position $b$. The CWT representation possesses axes for time, frequency, and the magnitude of the CWTs, which can then be placed on the third axis, thus creating a visual representation using the wavelet transform in a three-dimensional space. (Zhao et al., 2022) Subsequently, a colorization scheme is employed to depict the CWT magnitudes within a two-dimensional image.

For each LSS test, the data recorded by the sensor measuring the moment on the steering column is converted into scalograms. The extraction of scalograms from the time series data was performed using MATLAB®. To make events visually comparable in the scalograms, the length of the scalogram images is adjusted according to the length of the measurement. Specifically, every 10 millisecond of measurement in the time series data is converted into 1.024 pixels in the scalogram and, therefore, the length of the scalograms has a direct relationship with the length of the measurements. To calculate CWT magnitudes from the recorded data, wavelets with frequencies ranging from 0.0272 Hz to 6.951 Hz were used. In the resultant CWT figures, smaller frequencies are placed at the bottom and higher frequencies at the top, and the frequency axis uses logarithmic scale. Examples of resulting scalograms are shown in Fig. 3.

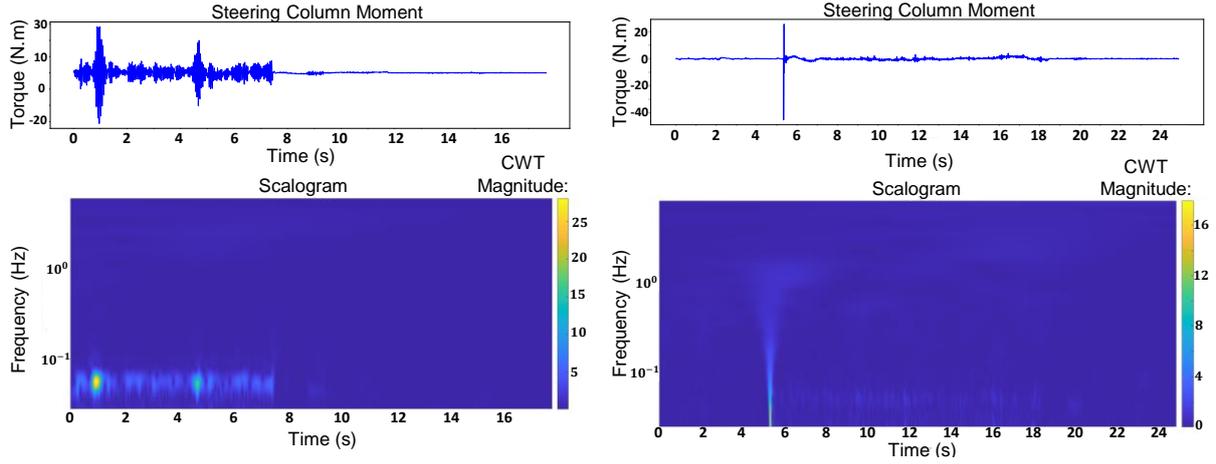

*Fig. 3. Two examples for time series data of the moment on the steering column measured throughout LSS tests (top) and their corresponding scalograms (bottom)*

### 4.1.2 Filtering

In the scalograms generated from the recorded time series data of the steering column moment, the objects representing the LDW event are vague and hardly visible. Since the time series data gathered from the AVS test are recorded in real-world and complex driving conditions, they contain other signals and numerous noisy effects which conceal the targeted events in the scalograms. Consequently, the accuracy of object detection networks for identifying those objects in the scalogram images decreases. To enhance the tracking process, it is therefore necessary to define a filter that highlights such events in the scalograms as a preprocessing step. At this point, some domain-specific knowledge about the CWT magnitude at the event spots in scalograms is required. By inspecting the CWT magnitudes at the locations in the scalograms where the steering wheel vibrates in response to the LDW effect, a tentative range for CWT magnitudes at these locations can be ascertained. Let $C_1$ and $C_2$ represent the minimum and maximum values of the CWT magnitudes of LDW spots, respectively. The interval ($C_1$, $C_2$) should be chosen as small as possible yet large enough to cover every possible CWT magnitude of LDW spots in the scalograms. Now, a filter is defined as follows: CWT magnitudes below the minimum value of the interval are set to zero, and the corresponding region in the scalogram are thereby displayed in white; for all CWT magnitudes that exceeded the maximum value of the specified interval, the magnitudes are set to the upper endpoint of the interval. Equation (3) provides the mathematical definition of the filter and Figure 4 illustrates the effect on the scalograms.

$$\begin{aligned} &if\ CWT < C_1 \rightarrow CWT = 0 \\ &if\ CWT > C_2 \rightarrow CWT = C_2 \end{aligned} \quad (3)$$

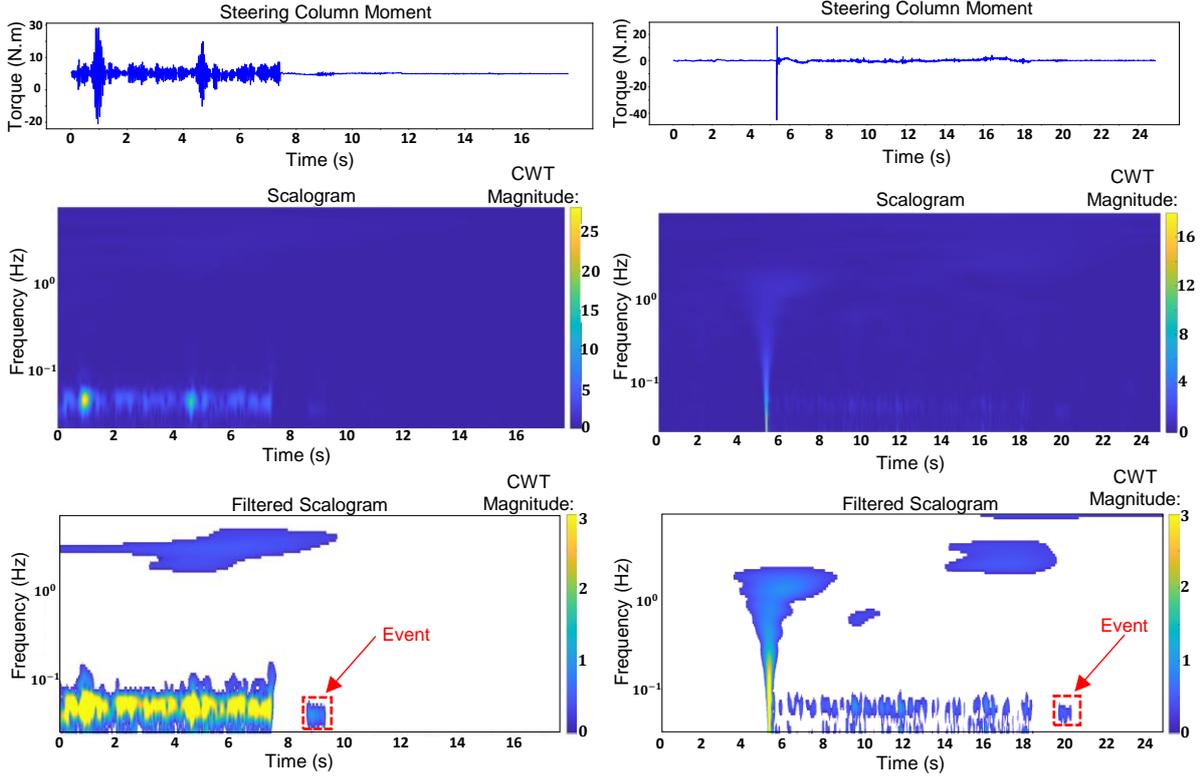

*Fig. 4. The impact of filtering and whitening on the scalograms.*

As a result of applying this filter, the CWT magnitudes which lie out of the determined interval disappear from the scalograms and only the magnitudes that fall within the interval stand out in a distinguishable manner with fully clear boundaries compared to the rest of the scalogram, paving the way for object detection. In the next step, an object detection algorithm is needed in order to find the objects representing the occurrence of the LDW event in the scalograms. We employ the YOLO model as described in the next section.

### 4.2 Object Detection in Scalograms

Object detection and localization algorithms play a crucial role in intelligent surveillance of the systems. (Jiang et al., 2022) Localization means determining the positions of the objects with a rectangular box around the object, called a bounding box. Generally, there are two types of object detectors: single-stage and double-stage detectors. Double-stage detectors initially use Region Proposal Networks (RPNs) to find Regions of Interest (RoIs) and subsequently identify the objects within those regions. In contrast, single-stage algorithms take the entire spatial area to detect possible objects in consideration and output bounding boxes along with a confidence score (CS) indicating the probability of object existence within a given region. Commonly, the accuracy of double stage detectors surpasses that of single stage detectors although single stage detectors have more efficient performances. Recently, You Only Look Once (YOLO) has become one of the most extensively used object detection networks in the field of computer vision. The precision of YOLO in detecting the objects is superior to that of two stage detectors. (Diwan et al., 2023) YOLO network treats the object detection task as a regression task and gets the bounding boxes and their CS at once. Being small in model size, high calculation speed and identifying the exact position of the detected object, are the main characteristics of YOLO algorithms. In this paper, we employ the YOLO version 5 which is briefly summarized in the following.

#### 4.2.1 YOLO Model and Network Architecture

Generally, YOLO Model divides the input image into $S \times S$ grids and then for each single grid cell, $B$ bounding boxes including their positions, dimensions, CSs and the probabilities for an object to belong to each of the predefined classes are predicted. (Xiao et al., 2022) As a result, the input image is encoded into a tensor with the size

$$S \times S \times [B \times (4 + 1 + C)] \tag{4}$$

where $B$ is the number of bounding boxes for each grid cell, $C$ is the number of classes, and four values store the coordinates of the bounding box and one value is used for the confidence score. For each bounding box, the YOLO model predicts a CS between (0,1) and only the predictions with a probability higher than a certain threshold are taken into consideration. Afterwards, using multilabel classification, the class is predicted for each bounding box; meanwhile, binary cross-entropy is used to compute the loss of class predictions. Then, the redundant bounding boxes are removed using non-maximum suppression (NMS), leaving the best matches to be identified as detection objects. According to the NMS principle, the algorithm selects the bounding box that has detected an object with the highest probability, namely the most confident bounding box and then, suppresses all the adjacent boxes that overlap the main bounding box. (Diwan et al., 2023) Single stage object detectors are mainly composed of three levels: Backbone, Neck and a Head to make dense predictions as shown in Fig. 5. The Backbone is a pretrained network which is used for extracting the features from images. The Neck of the model is designed for collecting the feature maps by generating a feature pyramids network and aggregating the features and passing them to the Head. The Head of the model carries out the final operations including predicting the CS for the bounding boxes and predicting the classes of the objects. (Wu et al., 2020)

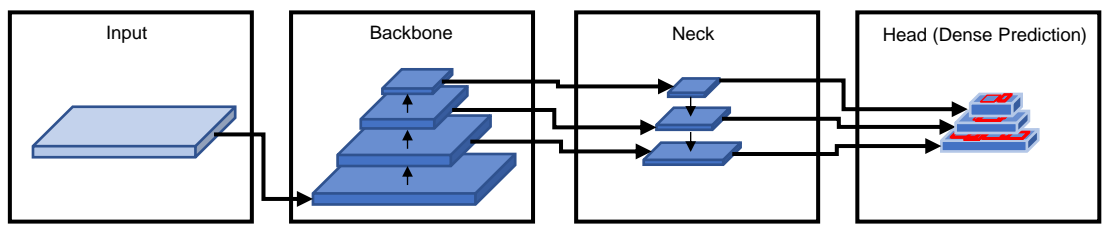

*Fig. 1. The architecture of single stage object detectors. (Wu et al., 2020)*

The architectural framework of YOLOv5 comprises several primary modules. The ConvBNSiLU is one of the basic modules of YOLOv5, composed of a convolution layer, a batch normalization layer, and a Sigmoid Linear Unit (SiLU) activation function. Additionally, the YOLOv5 model uses residual and dense blocks to transfer information to the deepest layers of the network. To avoid the problem of redundant gradients resulting from dense and residual blocks, the Bottleneck Cross Stage Partial (BottleneckCSP) module has been designed in YOLOv5 for feature extraction. (Wang et al., 2022) Furthermore, the Spatial Pyramid Pooling (SPP) network is another module used in YOLOv5, which extracts the most relevant features by aggregating information derived from the input and producing output of the same size without slowing down the network. (Qu et al., 2022) The complete architecture of YOLO v5 is shown in Fig. 6.

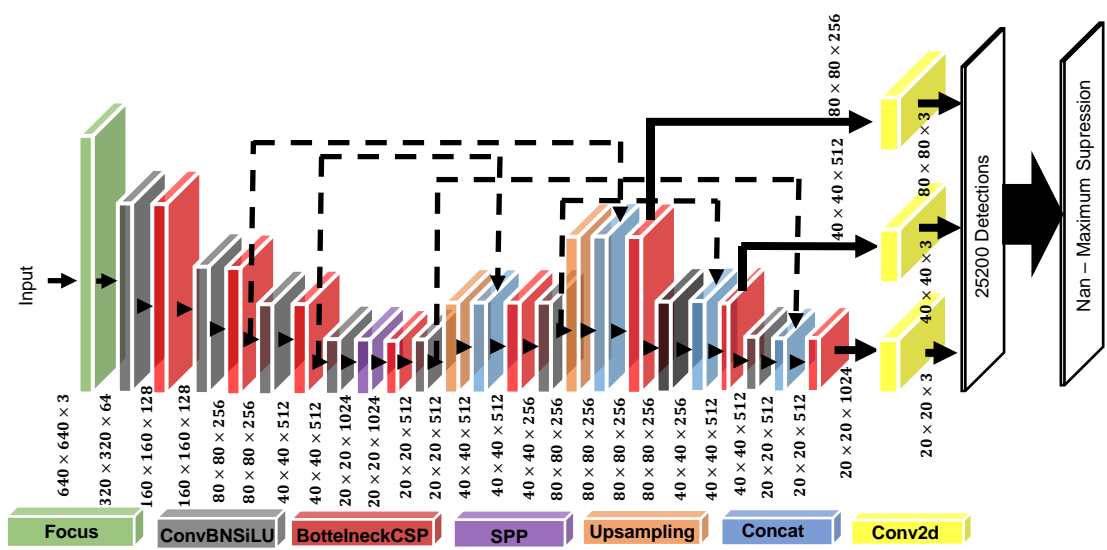

*Fig. 6. Architecture of YOLO V5 model. (Qu et al., 2022)*

### 4.2.2 Data Set and Labeling

Once the objects representing LDW events are made conspicuous and discernible in the scalograms via filtering, various samples of such objects are subsequently used to train the object detection model. To this end, the objects representing LDW events in the scalograms are identified based on an occurring event in the CAN-Bus data which serves as the ground truth. The labeling software LabelImg is used to capture and label these LDW objects. (Tzutalin, 2015) This software provides a user-friendly interface for drawing bounding boxes and assigning labels to objects within an image. LabelImg is commonly used in computer vision for creating training datasets for object detection or image classification tasks. The annotations are typically saved in XML format which can then be used for training ML models. (Tzutalin, 2015) The objects are annotated with rectangular bounding boxes encompassing them, thereby incorporating the coordinates of two points along the bounding box's diagonal into the labeled information. Fig. 7 depicts several instances of the entities that represent LDW events within the filtered scalograms.

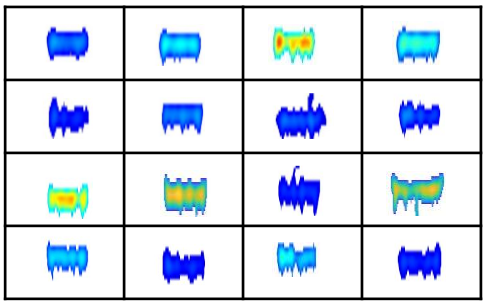

*Fig. 7. Examples of the objects representing the LDW events in the filtered scalograms*

A dataset consisting of the recorded data gathered from 518 LSS tests with overall 304 LDW event spots in the scalogram images was available for this study. Since the dataset contained the recorded time series data of different lengths, in order to avoid having extremely large size scalograms in the dataset and slowing down the detection as a result, the scalograms were split into image parts of at most 4,000 pixel width representing the time series data of 40 seconds in size and thereby discarding subparts lacking a vibration. The dataset is divided randomly into training, validation, and inference sets with proportions of 55%, 15%, 30%, respectively. The training and validation sets are used during the YOLO model training phase. The inference data is subsequently used to verify the final performance and generalization ability of the trained model for unseen samples. It is kept separate from the training and validation data to provide an unbiased verification and estimate how the model is likely to perform when deployed in real-world scenarios. (Xu and Goodacre, 2018)

### 4.2.3 Training Algorithm for the YOLO Model

The flow chart of the training algorithm is shown in Fig. 8. Given filtered and labeled samples for LDW events, the YOLO event detection model is trained to identify such events. In this study, three versions of the YOLOv5 model with different network sizes were trained and compared, namely YOLOv5n (Nano), YOLOv5s (Small) and YOLOv5m (Medium). The primary discriminant among these models lies in the variation in the quantity of feature extraction modules and convolution kernels they possess, which holds substantial practical relevance when working with YOLO models. Consequently, the parameter set size for YOLOv5n, YOLOv5s, and YOLOv5m varies, with values of 4 MB, 14 MB, and 41 MB, respectively. (Horvat et al., 2022) The object detection models were trained based on Python 3.10.5, PyTorch 1.13.0 used for YOLOv5. Since the desired objects have been highlighted by filtering, the depth of the YOLO architecture does not play a crucial role in the detection process. In general, filtering the scalograms simplifies and accelerates the learning processes of the YOLO network and greatly improves the system's reliability and accuracy.

The training data is used to train the YOLO model such that it learns patterns and relationships from this data. The validation data is used to fine-tune the model's hyperparameters and assess if it has learned the underlying patterns without overfitting. The relatively small sized training datasets compared to the large number of parameters in the YOLO model could lead to overfitting and therefore transfer learning should be utilized to train the model. Thus, pretrained YOLOv5 models were used as the model bases and the models were pretrained using the large-scale "COCO" dataset, a dataset used for object detection of over 300,000 images with 80 different object categories. (Ghiasi et al., 2018) The actual training for LDW events was performed according to the following hyperparameters: The models are trained for 150 epochs with a batch size of 16. For the other parameters, default values were used such that the momentum is set as 0.937 with a weight decay of 0.0005. The default SGD optimizer was used, and the training was performed with an initial learning rate of 0.01.

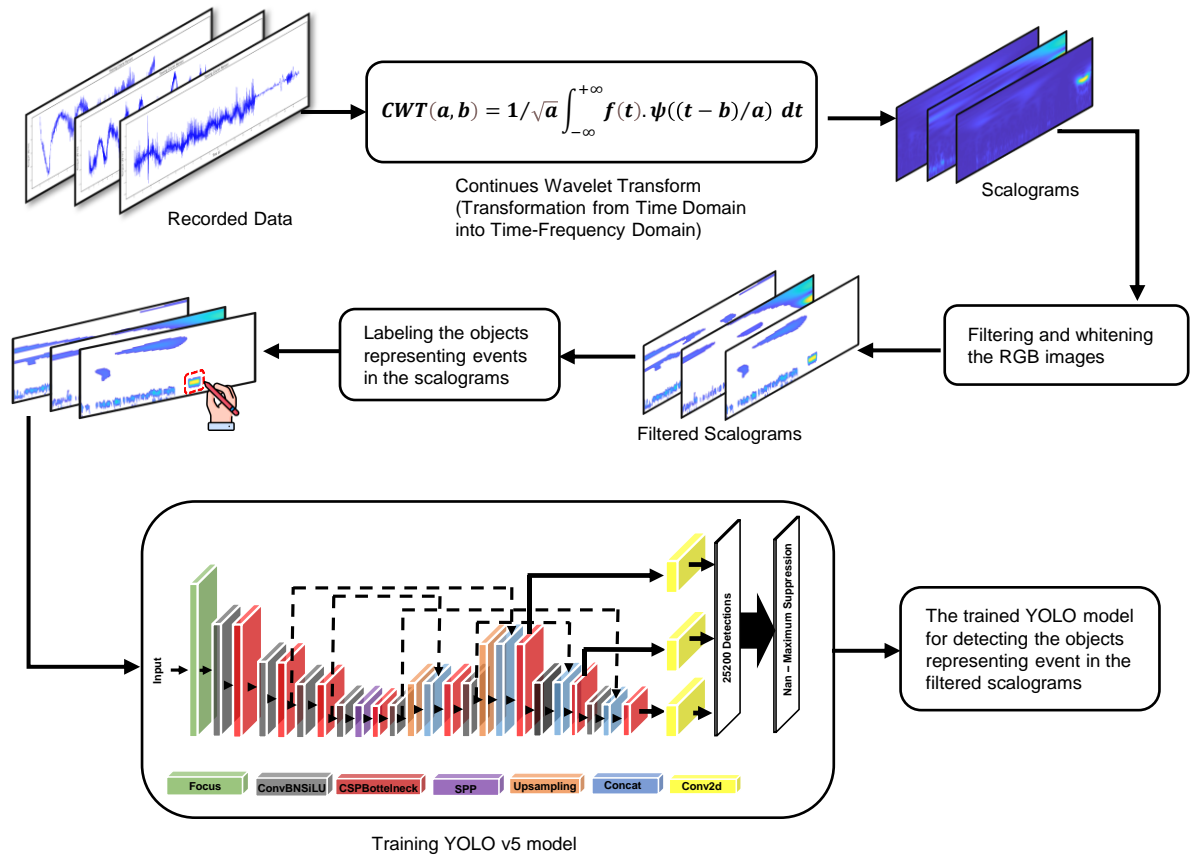

*Fig 8. Training Algorithm Flowchart*

To evaluate the performance of different versions of YOLOv5 models along the training iterations, in Fig. 9 the loss function, mean Average Precision (mAP), Precision and Recall are shown. The loss function in training a network refers to the error between the predicted output of the neural network and the actual output, and mAP is commonly used to evaluate the performance of object detection models by measuring the accuracy of the model in localizing and classifying objects in an image. (Dong et al., 2022) Figure 9.1 graphically depicts the dynamic progression of loss across epochs during the training procedure of the YOLOv5 network. Figure 9.2 visually presents the evolution of mean Average Precision (mAP) with respect to epochs during the comprehensive training process of the network. Figures 9.3 and 9.4 exhibit the precision and recall curves, elucidating the accuracy attained by the network at each epoch in correctly identifying the desired objects from the entire set of objects. The superior performance of the YOLOv5m version compared to YOLOv5n and YOLOv5s versions is evident in all four diagrams, particularly in Fig 9.3 and Fig 9.4. The YOLOv5m model achieved convergence at earlier epochs during training, which can potentially be attributed to the YOLOv5m network's higher parameter count, enabling it to effectively capture and represent intricate image attributes of greater complexity.(Horvat et al., 2022)

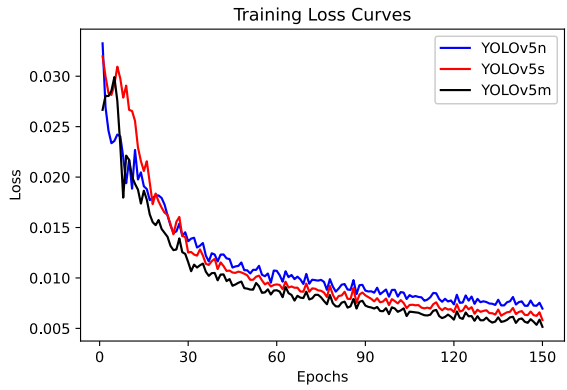

*Fig.9.1. Loss curves during training with a CS of 40%*

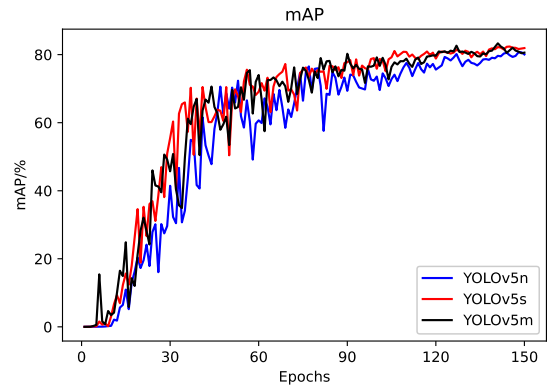

*Fig 9.2. mAP[0.5:.95] curves during training*

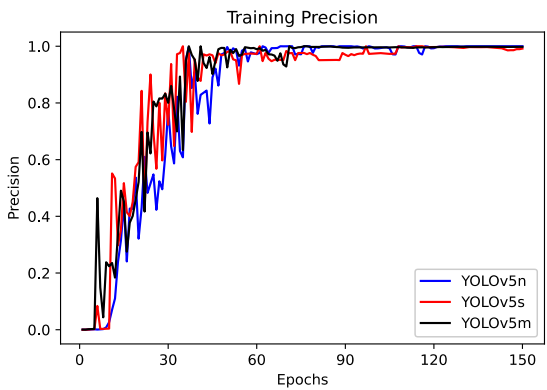

*Fig 9.3. Precision curves during training*

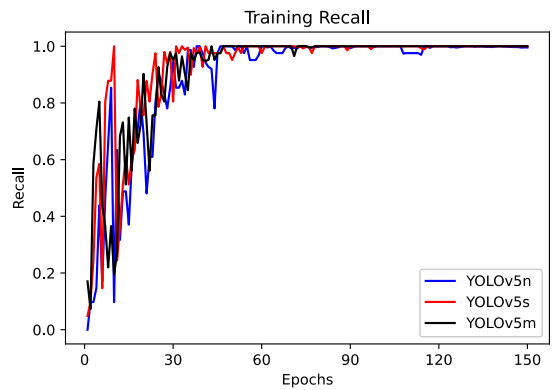

*Fig 9.4. Recall curves during training*

### 4.3 Frequency-Event Detection in Time Series Data

The proposed detection algorithm for frequency events can now be stated. The flow chart of the inference process is shown in Fig. 10. For a new, unseen time series data, a corresponding scalogram is generated and filtered. Thereafter, the object detection model seeks for the object representing the LDW event in the scalogram and if the intended object is detected with a probability higher than a specified CS, the corresponding time interval is projected onto the time series data, marking the detected time interval as event-occurring interval.

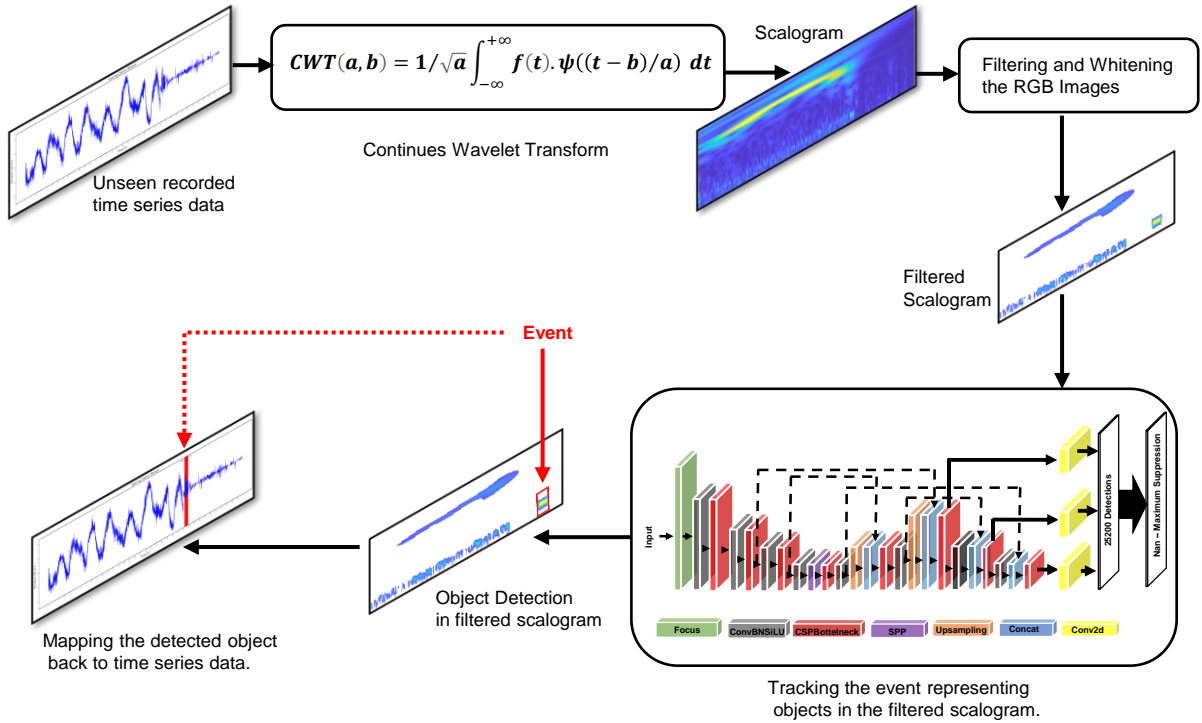

*Fig 10: Inference process to detect LDW events in time series data.*

## 5   Results and Discussion

The proposed algorithm is analyzed in detail in the following. The algorithm underwent verification for unseen LSS tests to ascertain its ability to detect the occurrences of LDW events in the time series data. To this end, the most reliable reference is the data captured by CAN-Bus sensors, which provide a distinct timeframe for the vibration detected on the steering wheel. This makes CAN-Bus sensors an optimal reference for evaluating the performance of the introduced algorithm.

Fig. 11 shows examples of detected LDW events in the time series data. It can be seen that the algorithm is capable to identify whether or not an LDW event has occurred in the presented moment data and, in addition, to indicate correctly the time interval of the event which closely corresponds to the CAN-Bus data. These capabilities are now studied quantitatively in the next sections.

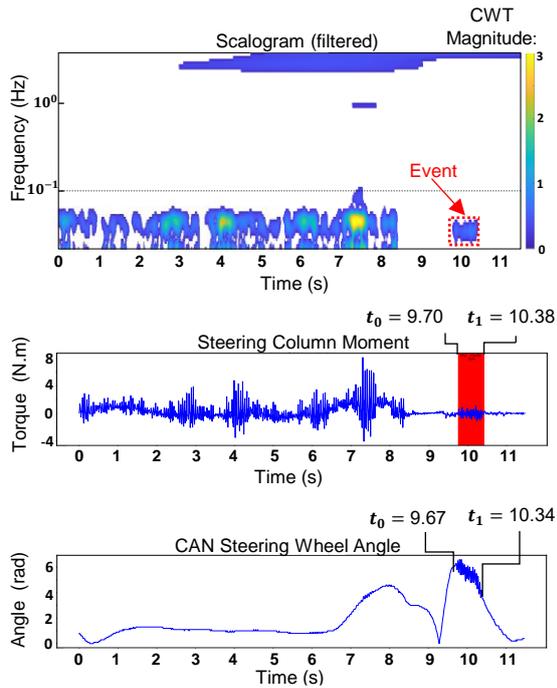

*Fig. 11 (a). The time interval of detected event by introduced algorithm: (9.70, 10.38). Time interval of vibrations sensed by CAN-Bus. (9.67, 10.34)*

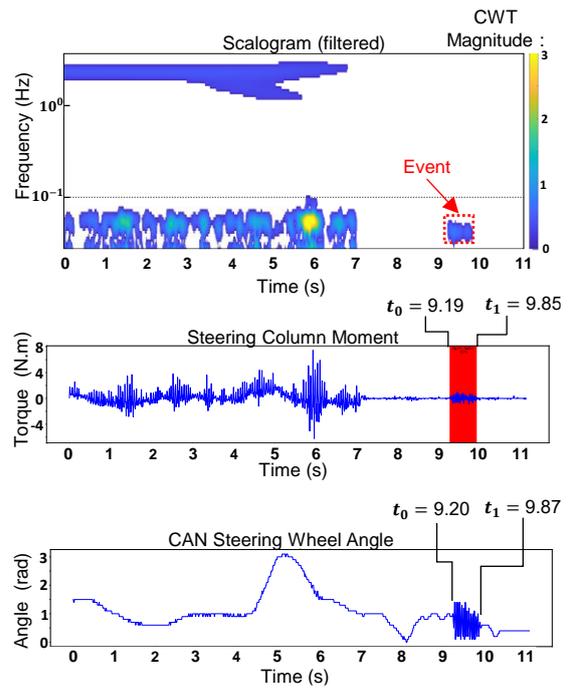

*Fig. 11 (d) The time interval of detected event using proposed algorithm: (9.19, 9.85) . Time interval of vibrations read by CAN-Bus. (9.20, 9.87)*

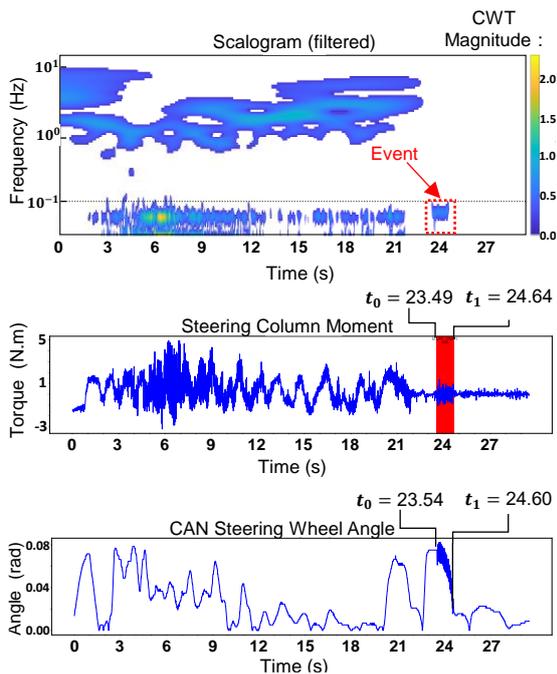

*Fig. 11(b). The time interval of detected event using proposed algorithm: (23.49, 24.64) . Time interval of vibrations read by CAN-Bus. (23.54, 24.60)*

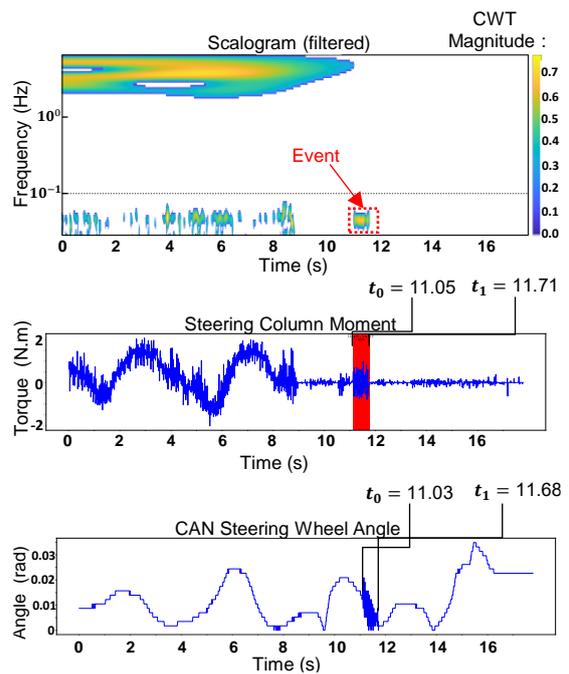

*Fig. 11 (e). The time interval of detected event using proposed algorithm: (11.05, 11.71) . Time interval of vibrations sensed by CAN-Bus (11.03, 11.68)*

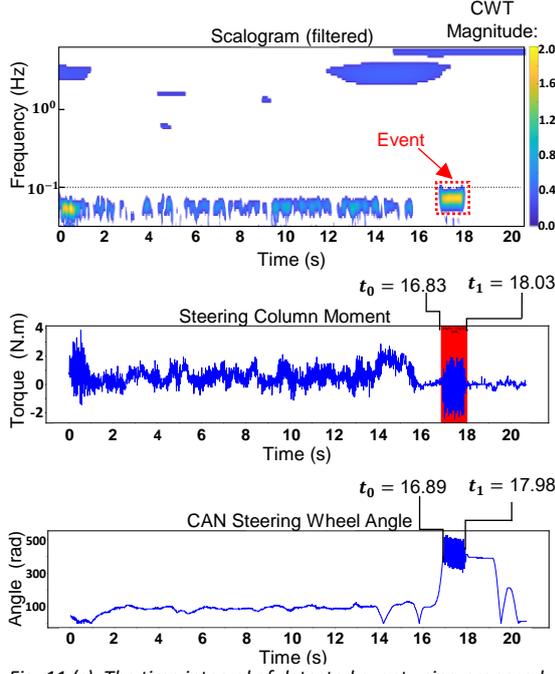
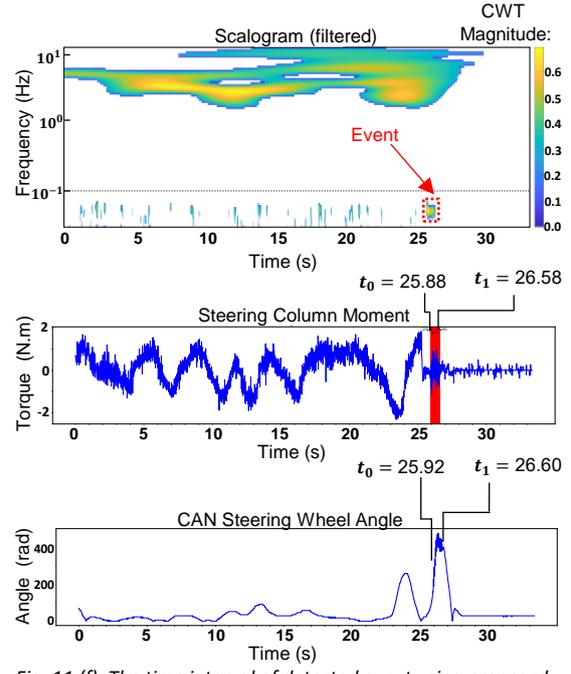

*Fig. 11 (c). The time interval of detected event using proposed algorithm: (16.83, 18.03) . Time interval of vibrations sensed by CAN-Bus (16.89, 17.98)*

*Fig. 11 (f). The time interval of detected event using proposed algorithm: (25.88, 26.58) . Time interval of vibrations sensed by CAN-Bus (25.92, 26.60)*

*Fig. 11. Detection of LDW events with wavelets of different ranges and the corresponding CAN-Bus data. Red bars on diagrams of steering column moment show the time interval of detected events using the proposed algorithm solely based on steering column moment data*

### 5.1 Performance of Frequency-Event Detection

To quantify the ability of the method to detect whether or not an LDW occurred in the time series data, the results on the inference data set are evaluated using the performance metrics Precision rate (P), Recall rate (R), Accuracy rate (ACC), and F1-score (F1), given by

$$P = TP/(FP + TP) \tag{7}$$
$$R = TP/(FN + TP) \tag{8}$$
$$ACC = (TP + TN)/(TP + TN + FP + FN) \tag{9}$$
$$F_1 = 2PR/(P + R) \tag{10}$$

True Positive (TP) is the number of events which are correctly detected by the algorithm. True Negative (TN) refers to the number of LSS tests in which no event has occurred throughout the test, and no event was detected by the algorithm. False Positive (FP) is the number of incorrect detections by the algorithm, and False Negative (FN) is the number of occurrences of the event throughout LSS tests which were not detected by the algorithm. Precision (P) indicates how many of the instances predicted as positive are actually positive by calculating the ratio of TP to the sum of TP and FP. Recall (R) is the metric, quantifying how well the model captures positive instances by the ratio of TP to the sum of TP and FN. Accuracy rate (ACC) refers to the ratio of correctly classified detections to all instances in the dataset, and finally, the F1-score is the harmonic mean of P and R. (Ma et al., 2019)

Table II provides an overview of the quantitative metrics using varying CS thresholds of 0.40, 0.60 and 0.75. In general, all three models demonstrate remarkable performance on the unseen dataset. However, it is apparent that the medium version of YOLO model outperforms the other models across all confidence thresholds, achieving an accuracy of 0.997 and an F1-score of 0.993 at a confidence threshold of 0.60 Nonetheless, this advantage comes at the expense of a slower inference speed and a greater overall model size. By increasing the CS threshold, the algorithm's efficacy in identifying objects indicative of LDW event is enhanced, as deduced from Table II. Nonetheless, employing an excessively high CS threshold may lead to the exclusion of objects that are correctly identified by the YOLO model. This observation is evident in the Table II, where elevating the CS from 0.60 to 0.75 results in a reduction in the values of multiple metrics. Consequently, the careful selection of an appropriate CS threshold is imperative to achieve optimal efficiency.

Table II. *Object detection results on the inference data set at different confidence thresholds.*

| Model | P | R | ACC | F1-score |
|---|---|---|---|---|
| *CS 40%:* | | | | |
| YOLOv5n (Nano) | 0.977 | 0.980 | 0.991 | 0.979 |
| YOLOv5s (Small) | 0.974 | 0.990 | 0.992 | 0.982 |
| YOLOv5m (Medium) | 0.974 | 0.997 | 0.994 | 0.985 |
| *CS 60%:* | | | | |
| YOLOv5n (Nano) | 0.980 | 0.970 | 0.989 | 0.975 |
| YOLOv5s (Small) | 0.983 | 0.967 | 0.989 | 0.975 |
| YOLOv5m (Medium) | 0.990 | 0.997 | 0.997 | 0.993 |
| *CS 75%:* | | | | |
| YOLOv5n (Nano) | 0.986 | 0.961 | 0.988 | 0.973 |
| YOLOv5s (Small) | 0.993 | 0.964 | 0.991 | 0.978 |
| YOLOv5m (Medium) | 0.990 | 0.993 | 0.996 | 0.992 |

### 5.2   Precision of Time Interval Identification

If an LDW event is detected together with its temporal interval by the algorithm, it can be compared to the ground truth indicated by the CAN-Bus data. Table III provides a comparison of the intervals obtained through both methods. The recorded data collected from the example LSS tests illustrated in Figures 11, 12, and 15 are instances utilized to juxtapose the LDW events detected by the algorithm with the vibrations derived from the CAN-Bus sensors in this paper. As shown in in Table III, the error rate in predicting the margin of the LDW event was generally below 10 percent in the majority of cases.

*Table III: A comparison of the intervals obtained by the introduced algorithm and the intervals sensed by the CAN-Bus sensors*

| LSS Tests shown in Fig: | LDW vibration indicated by | | Length of the LDW Event (seconds) | Prediction Error | | | |
|---|---|---|---|---|---|---|---|
| | the CAN-Bus sensors | the introduced algorithm | | in the beginning of the LDW time interval (seconds) | in the beginning of the LDW time interval. (%) | at the end of the LDW time interval (seconds) | at the end of the LDW time interval (%) |
| | $(\alpha_1, \beta_1)$ | $(\alpha_2, \beta_2)$ | $L = \beta_1 - \alpha_1$ | $\alpha_2 - \alpha_1$ | $|\alpha_2 - \alpha_1|/L$ | $\beta_2 - \beta_1$ | $|\beta_2 - \beta_1|/L$ |
| Fig. 11(a) | (9.67, 10.34) | (9.70, 10.38) | 0.67 | +0.03 | 4 | +0.04 | 6 |
| Fig. 11(b) | (23.54, 24.60) | (23.49, 24.64) | 0.67 | -0.01 | 1 | -0.02 | 2 |
| Fig. 11(c) | (16.89, 17.98) | (16.83, 18.03) | 1.06 | -0.05 | 4 | +0.04 | 3 |
| Fig. 11(d) | (9.20, 9.87) | (9.19, 9.85) | 0.65 | +0.02 | 3 | +0.03 | 4 |
| Fig. 11(e) | (11.03, 11.68) | (11.05, 11.71) | 1.06 | -0.06 | 5 | +0.05 | 4 |
| Fig. 11(f) | (25.92, 26.60) | (25.88, 26.58) | 0.68 | -0.04 | 6 | -0.02 | 3 |
| Fig. 12 (a) | (30.39, 31.45) | (30.35, 31.46) | 1.06 | -0.04 | 4 | +0.01 | 1 |
| Fig. 12 (b) | (35.44, 36.51) | (35.41, 36.56) | 1.07 | -0.03 | 3 | +0.05 | 5 |
| Fig. 12 (c) | (25.19, 25.86) | (25.15, 25.84) | 0.67 | -0.04 | 6 | -0.02 | 3 |
| Fig. 12(d) | (16.83, 17.50), unclear boundary on CAN-Bus | (16.80, 17.50), (20.49, 20.77) | 0.67, - | -0.03 | 4 | 0 | 0 |
| Fig. 12(e) | (16.64, 17.31) (20.56, 21.20) | (16.59, 17.31) (20.52, 21.21) | 0.67, 0.64 | -0.05, -0.04 | 7, 6 | 0,+ 0.01 | 1 |
| Fig. 12(f) | (14.22, 14.88) | (14.23, 14.81) | 0.70 | +0.01 | 1 | -0.07 | 10 |
| Fig. 12(g) | (18.52, 19.20) | (18.53, 19.18) | 0.68 | +0.01 | 1 | -0.02 | 3 |
| Fig. 12(h) | (14.70, 15.36) | (14.69, 15.33) | 0.66 | -0.01 | 1 | -0.03 | 4 |
| Fig. 15(a) | (11.03, 11.68) | (11.05, 11.71) | 0.65 | +0.02 | 5 | +0.03 | 8 |
| Fig. 15(b) | (12.90, 13.98) | (12.84, 14.01) | 1.08 | -0.06 | 5 | +0.03 | 2 |
| Fig. 15(c) | (10.09, 11.26) | (10.03, 11.26) | 1.10 | -0.06 | 5 | 0 | 0 |
| Fig. 15(d) | (14.45, 15.60) | (14.40, 15,61) | 1.10 | -0.05 | 4 | +0.01 | 1 |

Detecting the precise time interval in which an event occurs is of paramount importance for this study. The emergence of LDW vibrations in the recorded time series data has an immediate impact on the magnitudes of the CWTs, resulting in the sudden appearance of LDW events in the filtered scalograms. This enables the precise detection of the corresponding time intervals in the scalograms and subsequently in the time series data. Figure 12 presents some examples in which manually discerning the event spots on the time series data was challenging by visual inspection. Nevertheless, the algorithm allows for the detection of these event spots with precise temporal boundaries and by comparing the detections with the recorded time series data from the CAN-Bus, the accuracy of the detections is underscored, which is not attainable through human inspection.

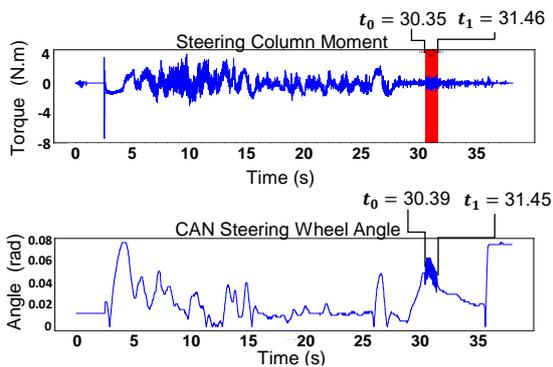

*Fig. 12(a). The time interval of detected event using algorithm: (30.35, 31.46). Time interval of vibrations sensed by CAN-Bus (30.39, 31.45).*

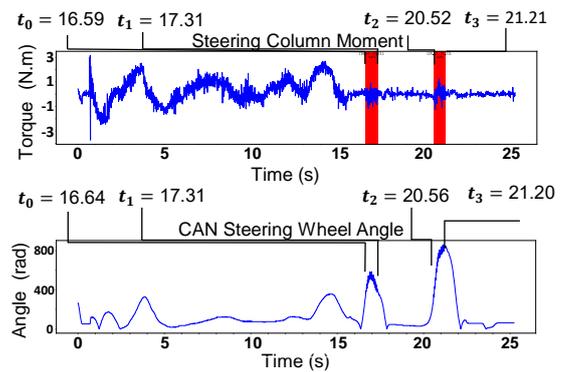

*Fig. 12(e). The time interval of detected event using algorithm: ( 16.59, 17.31 ) , ( 20.52, 21.21). Time interval of vibrations sensed by CAN-Bus (16.64, 17.31) (20.56, 21.20).*

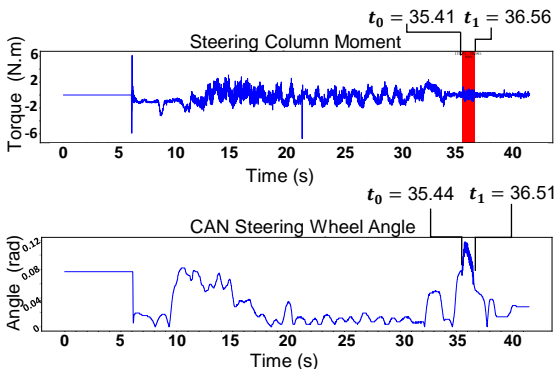

*Fig. 12(b). The time interval of detected event using algorithm: (35.41, 36.56). Time interval of vibrations sensed by CAN-Bus (35.44, 36.51).*

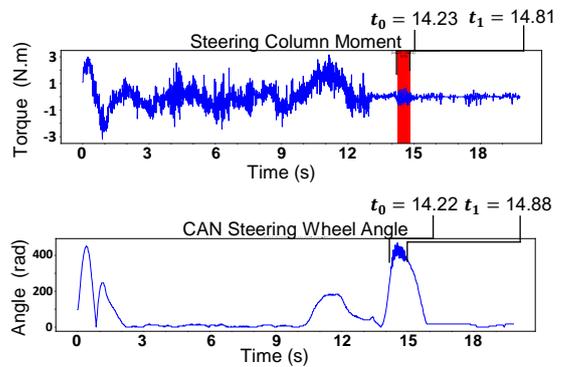

*Fig. 12(f). The time interval of detected event using algorithm: (14.23, 14.81 ). Time interval of vibrations sensed by CAN-Bus (14.22, 14.88).*

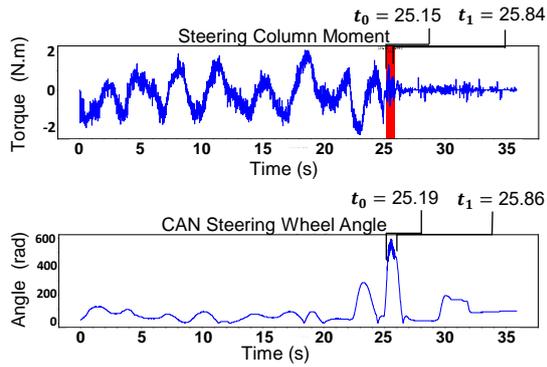

*Fig. 12(c). The time interval of detected event using algorithm: (25.15, 25.84 ). Time interval of vibrations sensed by CAN-Bus (25.19, 25.86).*

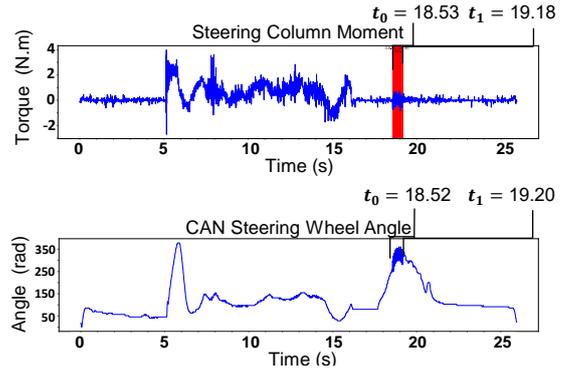

*Fig. 12(g). The time interval of detected event using algorithm: (18.53, 19.18 ). Time interval of vibrations sensed by CAN-Bus (18.52, 19.20).*

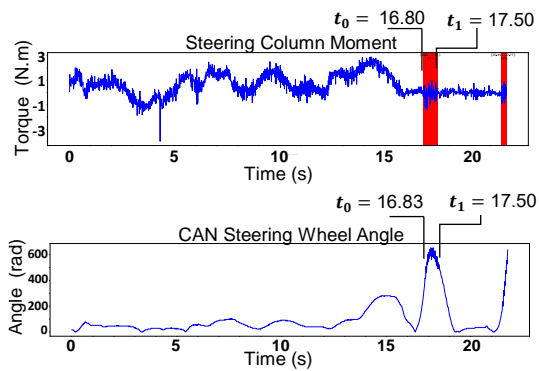

*Fig. 12(d). The time interval of detected event using algorithm: (16.80, 17.50 ) , (20.49 , 20.77 ) . Time interval of vibrations sensed by CAN-Bus (16.83, 17.50), unclear boundry on CAN-Bus.*

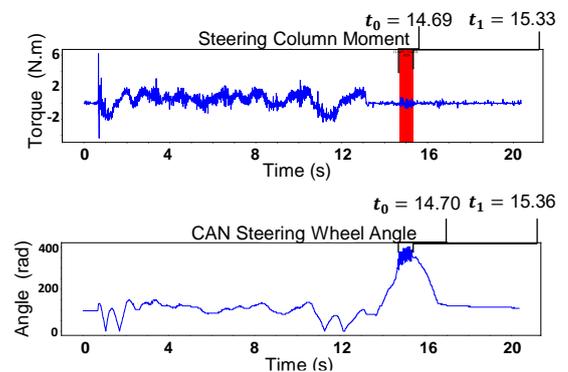

*Fig. 12(h). The time interval of detected event using algorithm: (14.69, 15.33) . Time interval of vibrations sensed by CAN-Bus (14.70, 15.36).*

*Fig. 12. Examples of detections which were challenging to track even manually by visual inspection.*

### 5.3   Incorrect or Failed Detections of the Algorithms

Despite the extremely convincing performance of the algorithm, in rare cases the method obtains FP or FN as indicated by the Precision and Recall scorings. In the following, the occurrence of such instances is studied in more detail. The simplicity of the objects representing LDW events poses a challenge, as vibrations similar to LDW effects in the steering column can generate objects with identical shapes in the filtered scalograms. These objects may be erroneously detected as LDW events by the YOLO model, resulting in FP results throughout the evaluation process. Fig. 13 presents examples of recoding in LSS tests where the appearance of a simple worm-shaped pattern in the scalogram has caused confusion for the model, leading to incorrect detections across all three versions of the YOLO model.

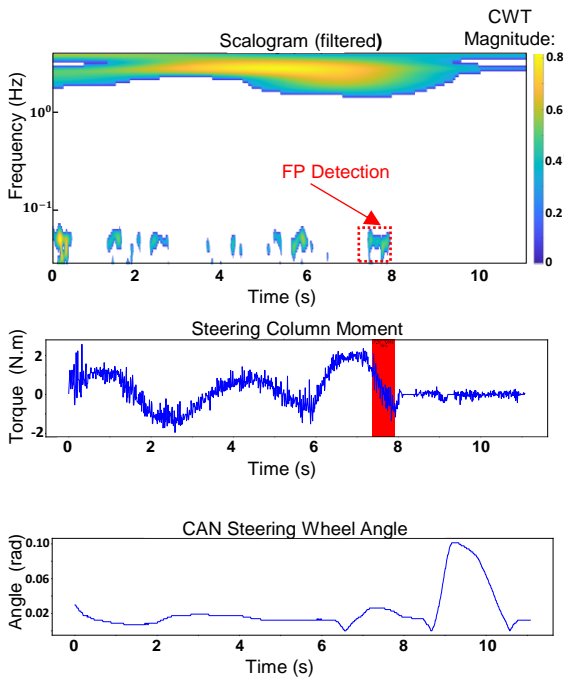
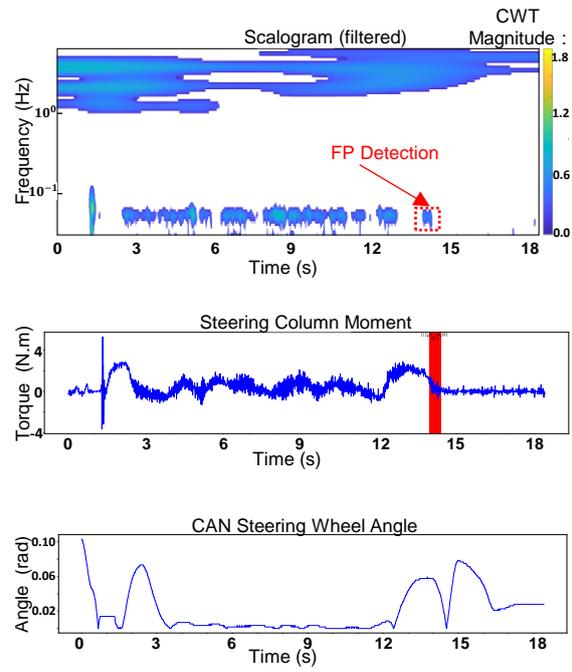

*Fig.13(a)*

*Fig.13(c)*

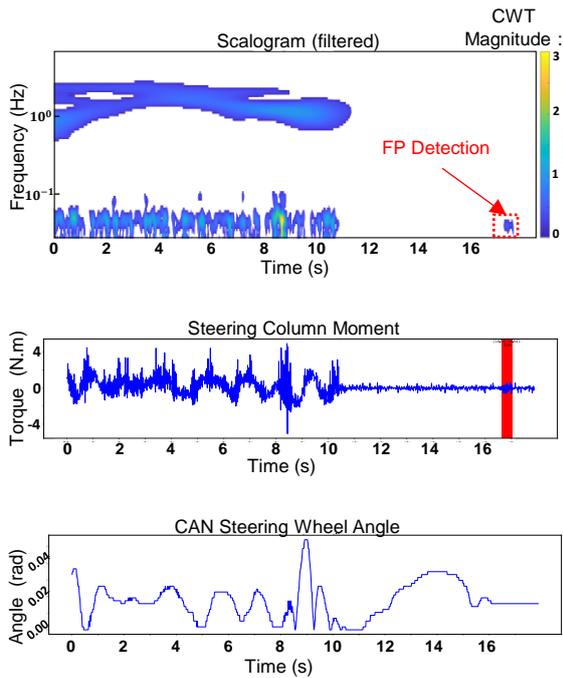

*Fig.13(b)*

*Fig. 13. Examples of LSS tests in which the emergence of a simple worm-shaped form in the scalogram has caused FN results.*

The precision of different versions of the YOLO model plays a pivotal role in identifying events in the scalograms. Fig. 14 presents several instances of FN cases, where the LDW entity went unnoticed by YOLOv5s but was discernible by either YOLOv5m or YOLOv5n. Typically, YOLOv5m exhibits higher accuracy than YOLOv5s and YOLOv5n since it encompass a greater number of parameters and can capture more complex image attributes. (Horvat et al., 2022) In Fig. 14(b), an LDW event is shown as detected by YOLOv5m, while it was not identified by either YOLOv5s or YOLOv5n shown in Fig. 14(a). Figure 14(d) presents another instance in which both

YOLOv5n and YOLOv5m were able to identify the LDW event, but YOLOv5s failed to detect it presented in Fig. 14(c). Notably, in Fig. 14(f), an LDW event is presented, which, despite its unusual appearance in filtered scalograms, was detected by YOLOv5n. However, neither YOLOv5m nor YOLOv5s detected it, as presented in Fig. 14(e).

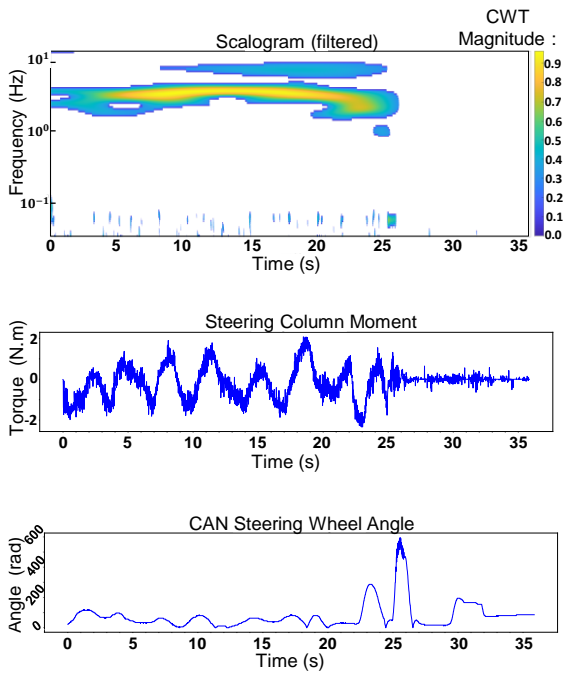

*Fig. 14(a). An example of LSS test where LDW event not identified by either YOLOv5s or YOLOv5n.*

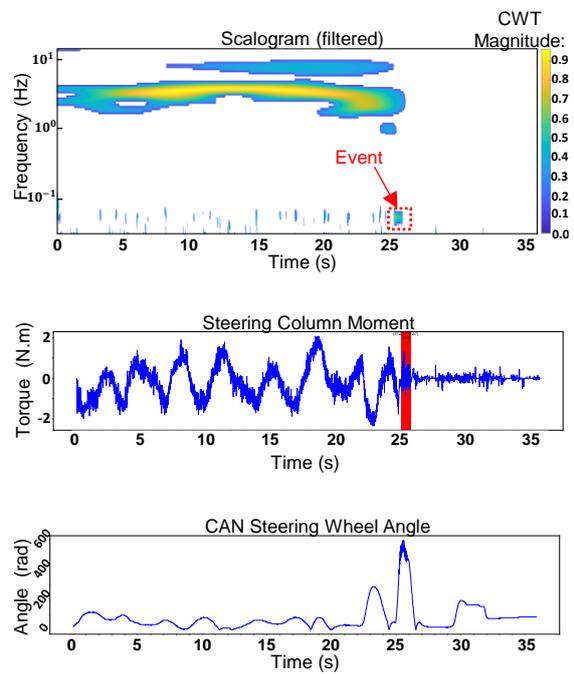

*Fig. 14(b). An example of LSS test where LDW event is shown as detected by YOLOv5m.*

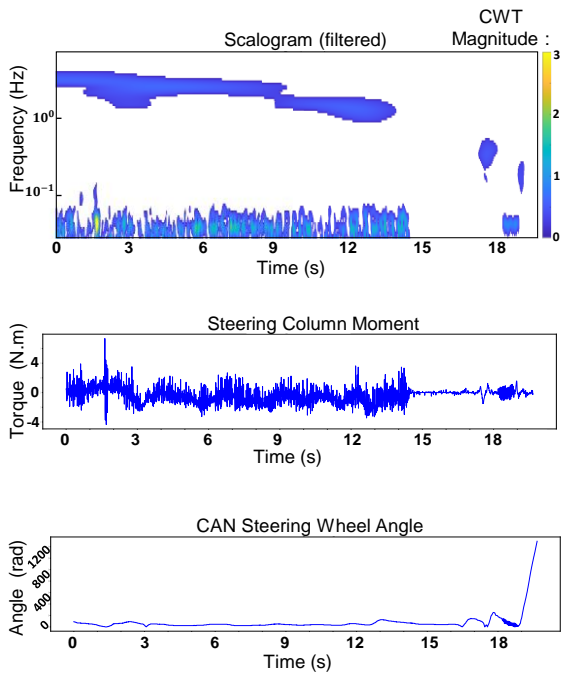

*Fig. 14(c). An example of LSS test in which YOLOv5s failed to identify the LDW event.*

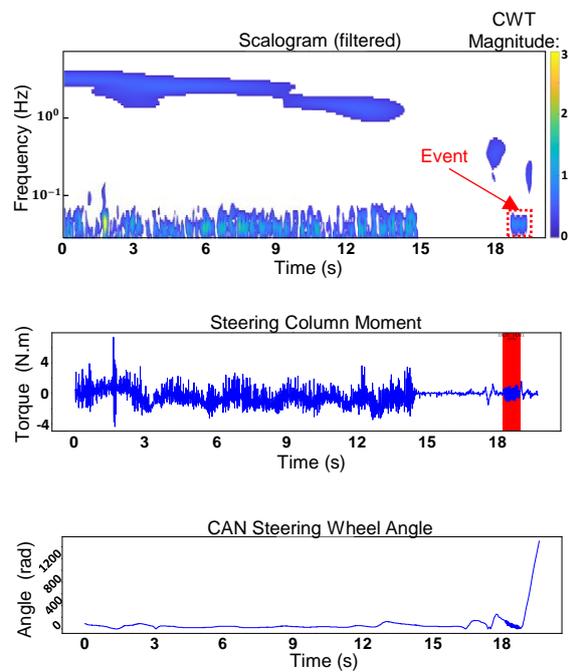

*Fig. 14(d). An example of LSS test in which both YOLOv5n and YOLOv5m were able to identify the LDW event.*

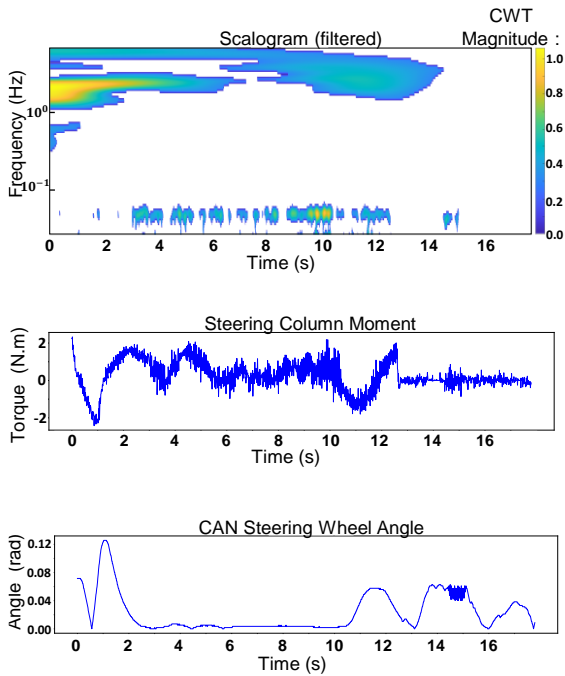 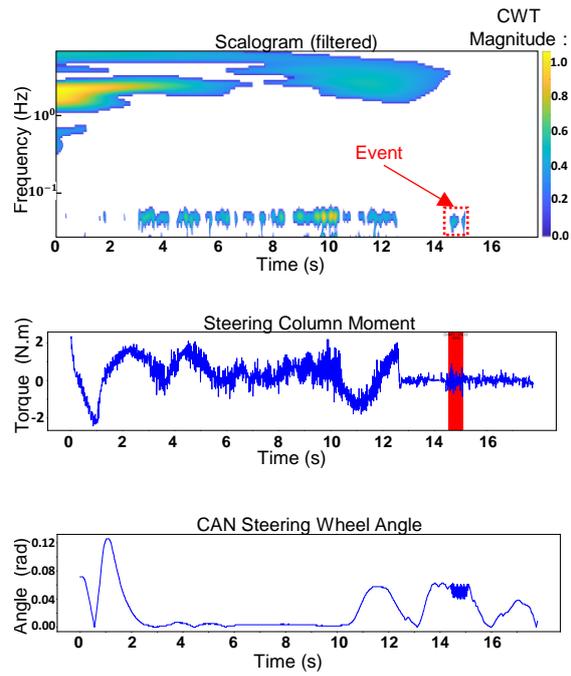

*Fig. 14(e). An example of LSS test where LDW event was not detected by YOLOv5s and YOLOv5m.*

*Fig. 14(f). An example of LSS test where LDW event was detected by YOLOv5n, despite its unusual appearance on filtered scalograms.*

*Fig. 14. The comparison between FN resulted which were fixed by choosing a different version of YOLO Model.*

## 5.4 Comparison with other Methodologies

The magnitude of the CWT at the LDW event spots exhibits substantial variation not only across different tests but also within a singular event throughout a single test. Moreover, upon examination of the recorded data obtained by the CAN-Bus sensor measuring the steering wheel angle, as depicted in Fig. 11, it becomes apparent that the steering wheel has undergone rotational and vibrational movements across significantly diverse angular ranges, implying that the LDW events have been recorded from various vehicles equipped with different types of steering wheels. Therefore, methodologies that rely on thresholds, ML-classifiers or frequency-based signal processing have proved inadequate in detecting the events in this study. Fig. 15 exemplifies such detections, whereby the non-uniform energy levels of the LDW vibration are discernible.

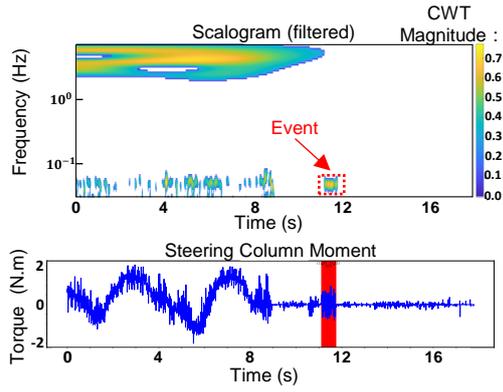
Fig. 15(a).

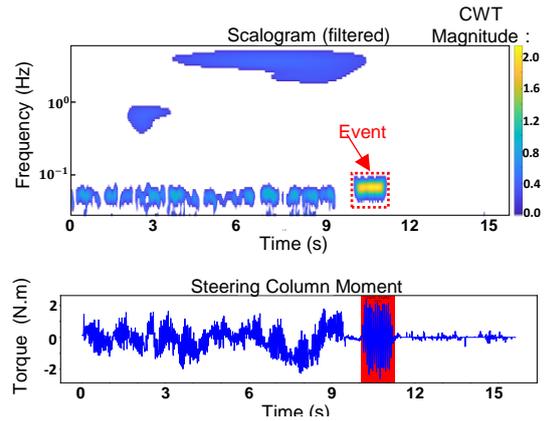
Fig. 15(c).

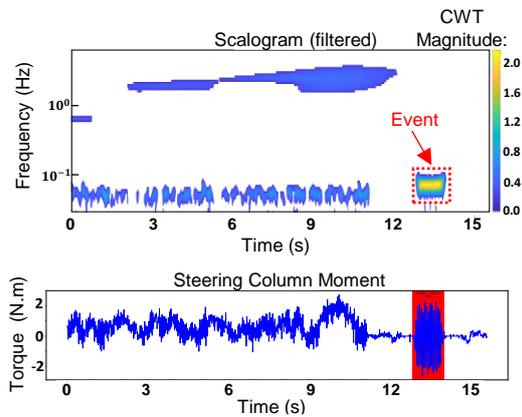
Fig. 15(b).

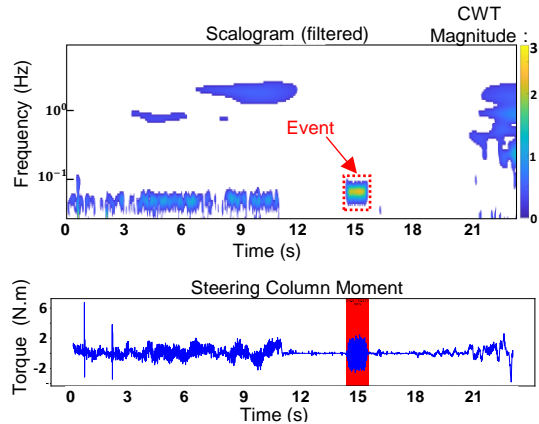
Fig. 15(d).

*Fig. 11. Magnitude of CWT in the event spots varies significantly, not only in different tests but also within a single event throughout one test.*

The relatively uniform distribution of the CWT magnitudes within the event spots has resulted in the emergence of LDW events as similar worm-shaped areas in the filtered scalograms. Therefore, the appearance of objects representing LDW events in the filtered scalograms in virtually identical forms is one of the most helpful clues in developing the methodology of this study. This yields an efficient event detection method through the use of object detection models. Upon comparing the objects representing LDW events across multiple scalograms, it became evident that the LDW events appeared at varying levels of the y-axis in the scalograms. This indicates that detection is accomplished using wavelets with different frequency ranges. As a result, frequency-based signal processing methods were once again deemed inadequate for providing proper indicators for event detection in this study. Examples of detections in different frequency ranges are shown in Fig. 11 with frequencies significantly less than $10^{-1}$ Hz as well as frequencies close to $10^{-1}$ Hz. However, as illustrated in Fig .11, the range of frequencies in which the LDW event's object is situated within the scalogram is limited. Therefore, this could be harnessed to restrict even more the frequency ranges in which the object is sought. This, in turn, might accelerate the convergence of the object detection network.

Generally, the presented algorithm offers a substantial advantage over ML-based classification methods and threshold-based techniques due to its capacity for generalization. In contrast to alternative approaches, the performance of this algorithm remains unaffected by certain influential factors, including the type of vehicle under examination, the measuring apparatus employed, and the driving conditions.

A potential shortcoming of our method arises when the steering wheel experiences pronounced vibrations during steering or when it undergoes rotational and vibrational motion with a frequency identical to that of the LDW  system. In such circumstances, there is a possibility that the vibrations generated by the LDW system may overlap and merge with the vibrations originating from the steering .Consequently, the vibrations emanating from the LDW system may become obscured both in the recorded time series data and in the scalograms. This could lead to misleading results in the detection process. However, within the scope of the real-world data available to us, we have not yet encountered instances of these phenomena. Nevertheless, we posit that even under such circumstances, it is plausible to differentiate between the diverse types of vibrations by examining their distinct frequency patterns that arise in the scalograms.

# 6 Conclusion

In this paper, a novel event detection method based on machine learning in the field of signal processing is proposed. The method transforms the time series data into a visual representation in the time-frequency domain via scalograms. To enhance the performance of the detection algorithm against imbalances and instabilities in the time-series data, filtering and modifying the scalograms have been employed in developing the proposed algorithm. After filtering the scalograms for highlighting the relevant event, a YOLOv5 model, as an object detection network, is employed to find the objects representing the desired events in the scalograms. Subsequently, the time intervals associated with the detected events are mapped back to the time series data to indicate the respective temporal spans. By employing the proposed algorithm, impressive performance metrics have been attained, including a precision rate of 0.97, a recall rate of 0.96, and an F1-score of 0.97 when detecting events within AVS data sets. Moreover, the algorithm accurately determines precise time boundaries for events, which can be difficult to ascertain through manual visual inspection. Hence, the implementation of this method will considerably enhance the accuracy and reliability of the vehicle testing analysis.

**CRediT authorship contribution statement**

**Bahareh Medghalchi:** Conceptualization, Methodology, Software, Validation, Formal analysis, Investigation, Writing - Original Draft, Writing - Review & Editing, Visualization.

**Andreas Vogel:** Conceptualization, Methodology, Writing - Review & Editing, Supervision.

**Declaration of Competing Interest**

The authors declare that they have no known competing financial interests or personal relationships that could have appeared to influence the work reported in this paper. The results, opinions and conclusions expressed in this publication are not necessarily those of Volkswagen Aktiengesellschaft.

**Appendix A. Supplementary material**

Due to the regulations of the company providing the data, survey respondents were assured raw data would remain confidential and would not be shared or published.